\documentclass[utf8,arxiv]{FrontiersinHarvard} %

\usepackage{url,hyperref,lineno,microtype,caption,subcaption, amsmath, graphicx, xcolor, tabularx, babel, pgfplots, algorithm, algpseudocode}
\usepackage[onehalfspacing]{setspace}

\def\keyFont{\fontsize{8}{11}\helveticabold }
\def\firstAuthorLast{von Arnim {et~al.}} 
\def\Authors{Axel von Arnim\,$^{1,2,*}$, Jules Lecomte\,$^{1,2}$,  Naima Elosegui Borras\,$^{3,4}$, Stanisław Woźniak\,$^{3}$, and Angeliki Pantazi$^{3}$}
\def\Address{$^{1}$fortiss GmbH, Neuromorphic Computing, Munich, 80805, Germany. \\
$^{2}$Equal contribution of the authors \\
$^{3}$IBM Research – Zurich, R{\"u}schlikon, 8803, Switzerland \\
$^{4}$University of Zurich and ETH Zurich, Zürich, 8006, Switzerland}
\def\corrAuthor{Axel von Arnim}

\def\corrEmail{vonarnim@fortiss.org}

\newcommand{\figArch}[1]{
\begin{figure}[h]
\centering
\includegraphics[width=\textwidth]{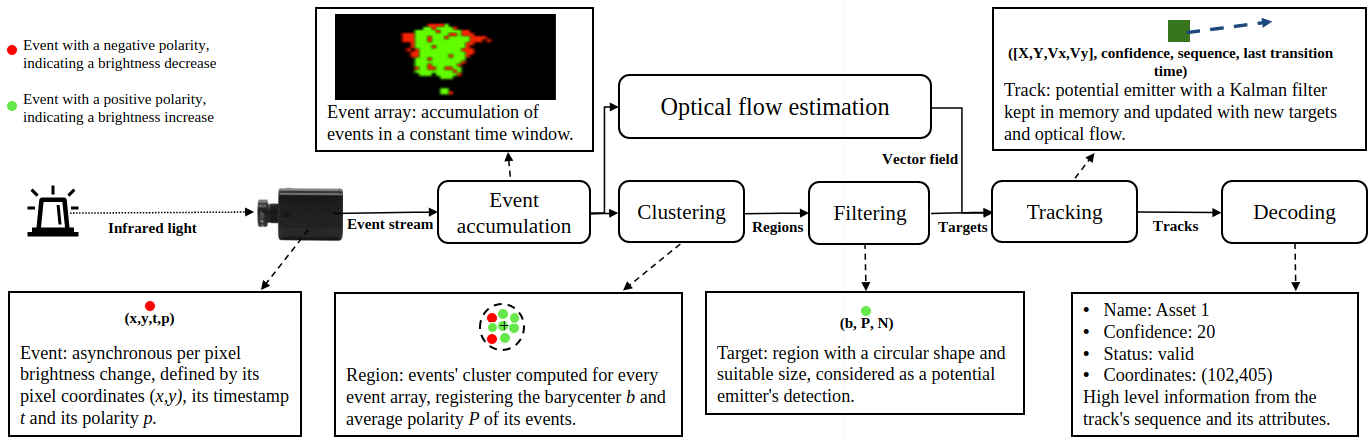}
\caption{\textbf{Architectural diagram of our system.} The beacon's light is detected by the sensor as events. Events are processed to track the beacons and further decode the transmitted messages. The event array block shows a snapshot of recorded events.}
\label{fig:architecture}
\end{figure}
}

\newcommand{\figSeqSum}[1]{
\begin{figure}[h]
\centering
\includegraphics[width=5cm]{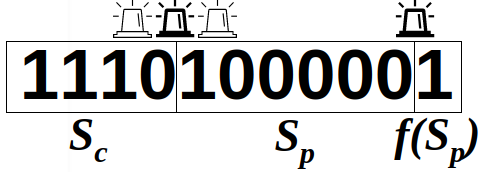} 
\caption{\textbf{A valid sequence}, decoded from blinking transitions.}
\label{fig:protocol}
\end{figure}
}

\newcommand{\figSNN}[1]{
\begin{figure}[h]
\centering
\includegraphics[width=0.5\textwidth]{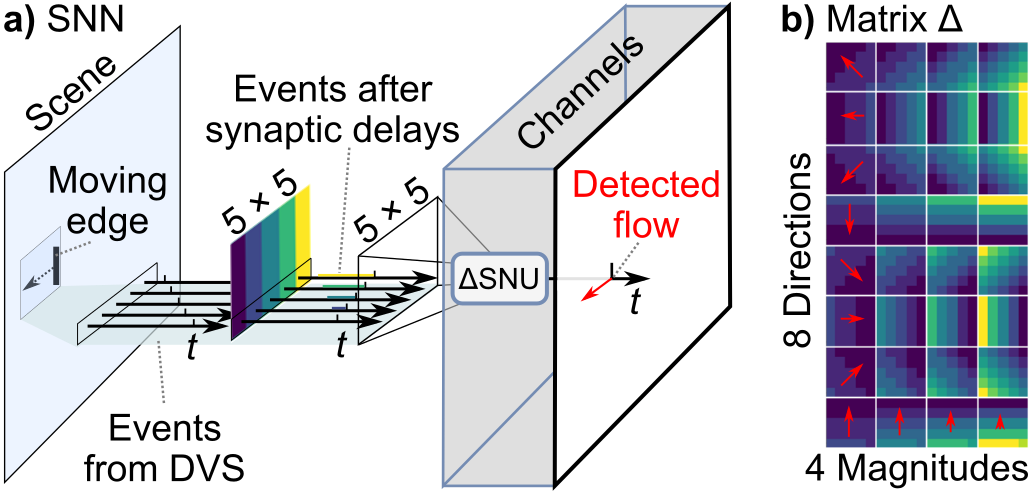} 
\caption{\textbf{SNN for sparse optical flow.} \textbf{a.} Events from camera at each input location are processed by 32 $\Delta$SNU units, each with specific synaptic delays. \textbf{b.} The magnitudes of synaptic delays are attuned to 8 different movement angles (spatial gradient of delays) and 4 different speeds (different magnitudes of delays), schematically indicated by red arrows.}
\label{fig:synapticdelays}
\end{figure}
}

\newcommand{\figTrac}[1]{
\begin{figure}[h]
\centering
\includegraphics[width=0.5\textwidth]{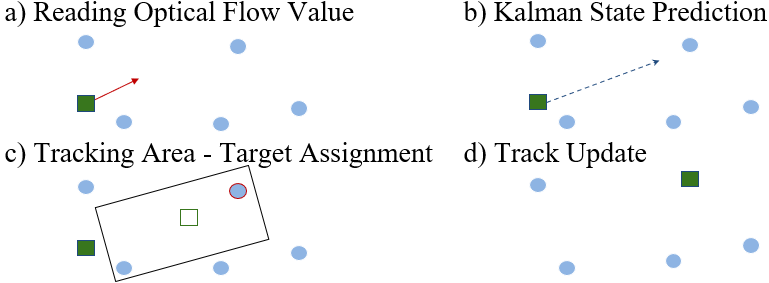} 
\caption{\textbf{Tracking steps.} \textbf{a}. Reading optical flow (red arrow) at the track's location. \textbf{b}. Prediction of the Kalman state via the track's location and the optical flow value. \textbf{c}. Tracks are assigned to a target in its oriented neighborhood, based upon the track's motion. \textbf{d}. The track's state, its size and its polarity are updated with the paired target's properties.}
\label{fig:tracking}
\end{figure}
}

\newcommand{\figLoops}[1]{
\begin{figure}[h]
\centering
\includegraphics[width=0.8\textwidth]{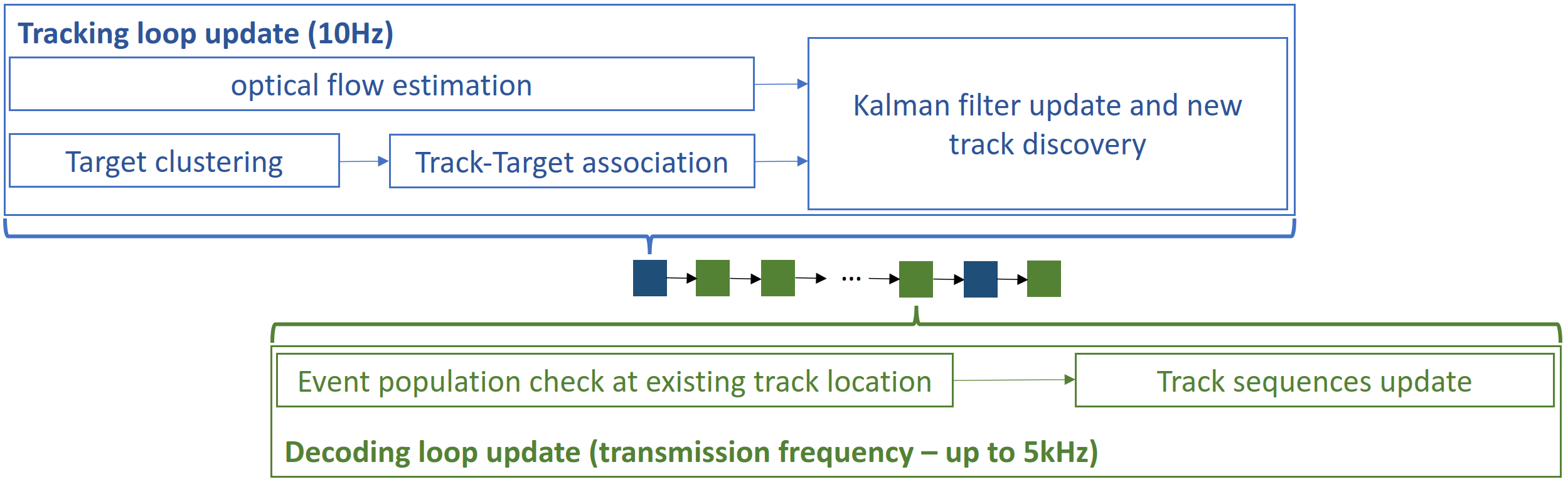} 
\caption{\textbf{Decoupled loops}: The tracking loop has a much lower and fixed frequency to maintain efficiency, while the decoding loop has the same frequency as the emitter to be able to decode the received signal.}
\label{fig:computation_reduction}
\end{figure}
}

\newcommand{\figSim}[1]{
\begin{figure}[h]
\centering
\includegraphics[width=0.5\textwidth]{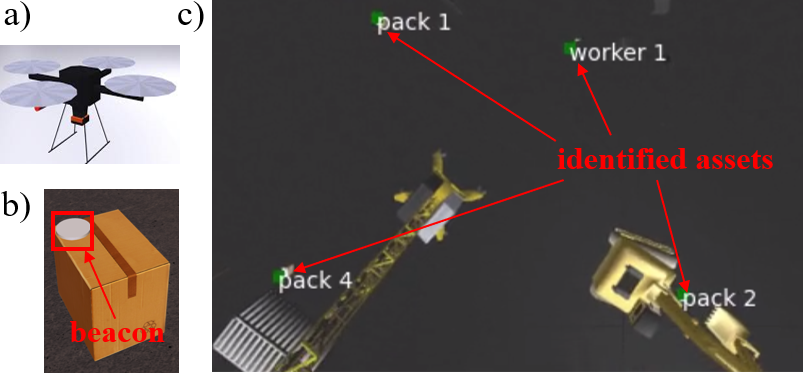} 
\caption{\textbf{Simulation setup.} \textbf{a}. Hector quadrotor. \textbf{b}. Example asset. \textbf{c}. The drone's point of view with decoding results.}
\label{fig:simu}
\end{figure}
}

\newcommand{\figResa}[1]{
\begin{figure}[h]
\centering
\includegraphics[width=0.8\textwidth]{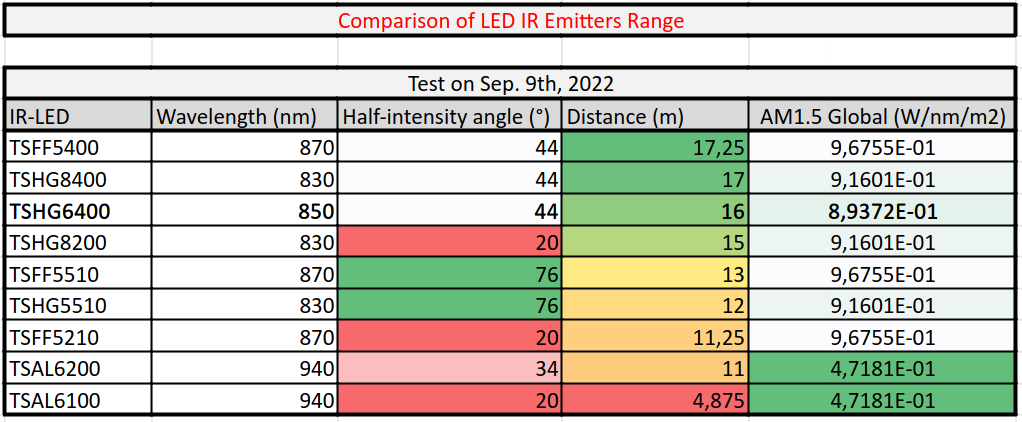} 
\caption{\textbf{LED wavelength benchmark}: The range of LEDs with varying wavelength and half-intensity angle was experimentally determined. AM1.5 Global is the solar integrated power density. The final choice of 850~nm ensures a good trade-off between detection range and outdoor solar irradiance.}
\label{fig:LED}
\end{figure}
}

\newcommand{\figled}[1]{
\begin{figure}[h]
\centering
\includegraphics[width=0.4\textwidth]{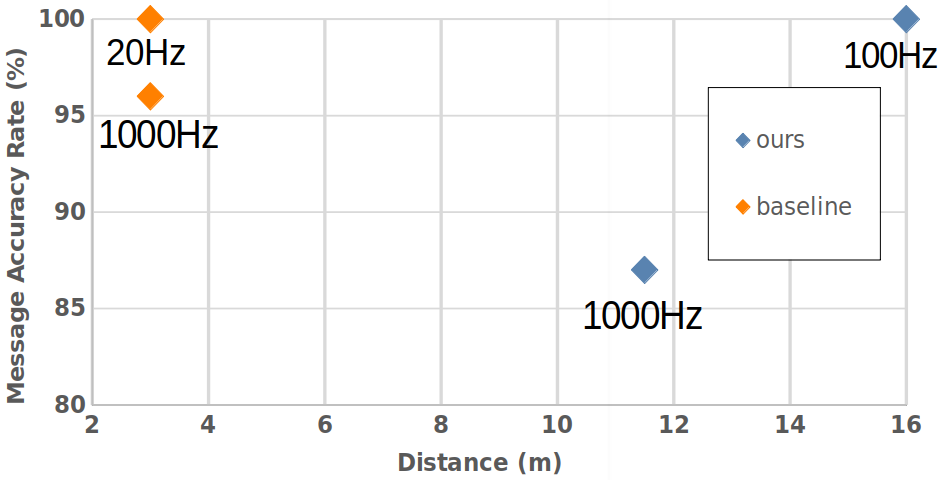} 
\caption{\textbf{Static OCC performance}: MAR for increasing beacon frequencies in comparison with the state-of-the-art baseline \citep{australianpaperocc}. Results were obtained at a 50~cm distance.}
\label{fig:hwresultsa}
\end{figure}
}

\newcommand{\figResb}[1]{
\begin{figure}[h]
\centering
\includegraphics[width=0.4\textwidth]{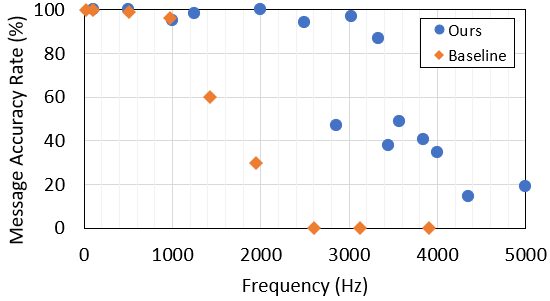} 
\caption{\textbf{Static OCC performance}: MAR for increasing beacon distance to the camera in comparison with the state-of-the-art baseline \citep{australianpaperocc}.}
\label{fig:hwresultsb}
\end{figure}
}

\newcommand{\tableMethods}[1]{
\begin{table}[#1]
\begin{center}
\begin{tabular}{|l|c|c|c|c|} 
    \hline
  \textbf{Method} & Type of camera & Data throughput (bps) & Tracking\\ 
  \hline \hline
  \citep{IV2007}  & frame-based & 250 & Yes\\ 
  \hline
  \citep{binningproblemeventbased} & event-based & 500 & No\\ 
  \hline
  \citep{australianpaperocc}& event-based & 500& No\\ 
  \hline 
  \citep{ScaramuzzaALMlocalisation} & event-based & identification only & Yes\\ 
  \hline
  \textbf{Ours} & event-based & 2500 & Yes\\ 
  \hline
\end{tabular}
\end{center}
    \caption{\textbf{Characteristics of existing identification methods}. The table presents existing optical camera communication solutions, using frame- or event-based cameras.}
    \label{tab:methods}
\end{table}
}

\newcommand{\tableResults}[1]{
\begin{table}[#1]
\begin{center}
\begin{tabular}{|l|c|c|c|c|} 
    \hline
  \textbf{Setup} & \textbf{Rate} & \textbf{Range} & \textbf{MAR} & \textbf{BAR}\\ 
  \hline \hline
  Simulation with optical flow  & 50~Hz & \textbf{28~m} & \textbf{74~\%} & 75~\%\\ 
  \hline
  Simulation without optical flow & 50~Hz & 28~m & 71~\% & 72~\%\\ 
  \hline
  Hardware & 2000~Hz & 5~m & 65~\% & \textbf{94~\%}\\ 
  \hline 
   Hardware & 2000~Hz & \textbf{16~m} & 27~\% & 87~\%\\ 
  \hline 
\end{tabular}
\end{center}
    \caption{\textbf{Identification performance with moving beacons}}
    \label{tab:comparison}
\end{table}
}

\newcommand{\figAlgorithm}[1]{
\begin{algorithm}[h]
\caption{Track classification with a confidence system}
\label{alg:confidence}
\begin{algorithmic}
\If{valid sequence} 
    \State confidence $\gets$ confidence+2
\Else
    \State confidence $\gets$ confidence-1
\EndIf 

\If{confidence $\geq$ confidence$_{\text{max}}$}
    \State status = valid
\ElsIf {confidence $\leq$ 0}
        \State status = invalid
\EndIf
\If{confidence $\leq$ confidence$_{\text{min}}$ or $t_c-t_t > $delay$_{\text{max}}$}
    \State forget track
\EndIf
\end{algorithmic}
\end{algorithm}
}

\begin{document}
\onecolumn
\firstpage{1}

\title {Dynamic Event-based Optical Identification and Communication} 

\author[\firstAuthorLast ]{\Authors} 
\address{} 
\correspondance{} 

\extraAuth{}

\maketitle

\begin{abstract}
Optical identification is often done with spatial or temporal visual pattern recognition and localization. Temporal pattern recognition, depending on the technology, involves a trade-off between communication frequency, range and accurate tracking. We propose a solution with light-emitting beacons that improves this trade-off by exploiting fast event-based cameras and, for tracking, sparse neuromorphic optical flow computed with spiking neurons. The system is embedded in a simulated drone and evaluated in an asset monitoring use case. It is robust to relative movements and enables simultaneous communication with, and tracking of, multiple moving beacons. Finally, in a hardware lab prototype, we demonstrate for the first time beacon tracking performed simultaneously with state-of-the-art frequency communication in the kHz range.



\tiny
 \keyFont{ \section{Keywords:} Neuromorphic Computing, Event-Based Sensing, Optical Camera Communication, Optical Flow, Identification}
\end{abstract}

\section{Introduction}

Identifying and tracking objects in a visual scene has many applications in sports analysis, swarm robotics, urban traffic, smart cities and asset monitoring. Wireless solutions have been widely used for object identification, such as RFID \citep{rfid} or more recently Ultra Wide Band \citep{UWB}, but these do not provide direct localization and require meshes of anchors and additional processing. One efficient solution is to use a camera to detect specific visual patterns attached to the objects.

This optical identification is commonly implemented with frame-based cameras, either by recognizing a spatial pattern in each single image -- for instance for license plate recognition \citep{licenseplate} -- or by reading a temporal pattern from an image sequence \citep{IV2007}. The latter is resolution-independent, since the signal can be reduced to a spot of light, enabling for much faster frame frequencies. It can be implemented with near-infrared blinking beacons that encode a number in binary format, similarly to Morse code, to identify assets like cars or road signs. But frame-based cameras, even at low resolutions, impose a hard limit on the beacon's frequency (in the $10^2$~Hz order of magnitude).
This technique is known as Optical Camera Communication (OCC) and has been developed primarily for communication between static objects \citep{OCC}.

Identifying static objects is possible with OCC as discussed before, but in applications such as asset monitoring on a construction site, it is also important to track dynamically moving objects. OCC techniques potentially enable simultaneous communication with, and tracking of, beacons. However, two challenges arise in the presence of relative movements: filtering out the noise and tracking the beacons' positions. Increasing the temporal frequency of the transmitted signal, since noise has lower frequencies than the beacon's signal, addresses this problem. Nevertheless, current industrial cameras do not offer a satisfying spatio-temporal resolution trade-off. Biologically-inspired event cameras, operating with temporally and spatially sparse events, achieve pixel frequencies on the order of $10^4$~Hz and can be combined with Spiking Neural Networks (SNNs) to build low-latency neuromorphic solutions.
They capture individual pixel intensity changes extremely fast rather than full frames \citep{binningproblemeventbased}. Early work combined the fine temporal and spatial resolution of an event camera with blinking LEDs at different frequencies to perform visual odometry \citep{ScaramuzzaALMlocalisation}. Recent work makes use of these cameras to implement OCC with smart beacons and transmit a message with the UART protocol \citep{australianpaperocc}, delivering error-free messages of static beacons at up to 4~kbps indoors and up to 500~bps at 100~m distance outdoors with brighter beacons, but without tracking. This paper, combined with the tracking approach presented in \citep{IV2007}, are the baseline of our work. The Table \ref{tab:methods} summarizes the properties of the mentioned methods.

\tableMethods{}
On the tracking front -- to track moving beacons in our case -- a widely used technique is optical flow \citep{kalmanfilteropticalflowtracking}. Model-free techniques relying on event cameras for object detection have been implemented  \citep{pedestriandetection_eventbased,tracking_event_based}.
To handle the temporal and spatial sparsity of an event camera, a state-of-the-art deep learning frame-based approach \citep{teed_raft_2020} was adapted to produce dense optical flow estimates from events \citep{gehrig_e-raft_2021}. However, a much simpler and more efficient solution is to compute sparse optical flow with inherently sparse biologically-inspired SNNs \citep{orchard_spiking_2013}, also considering network optimisation and improved accuracy \citep{schnider_2023}.

In this paper, we propose to exploit the fine temporal and spatial resolution of event cameras to tackle the challenge of simultaneous OCC and tracking, where the latter is based on the optical flow computed from events by an SNN. We evaluate our approach with a simulated drone that is monitoring assets on a construction site.
We further introduce a hardware prototype comprising a beacon and an event camera, which we use for demonstrating an improvement over state-of-the-art OCC range. To our knowledge, there is no method combining event-based OCC with tracking to identify moving targets. Furthermore, we beat the transmission frequency of our baseline.

\section{Materials and Methods}

The system that we propose is composed of an emitter and a receiver. The former is a beacon emitting a temporal pattern (a bit sequence) with near infrared light (visible to cameras, but not to humans), attached to the object to be identified and tracked. The receiver component is an event-based camera connected to a computer which, in turn, executes decoding and tracking algorithms. The receiver part comprises algorithmic components for clustering and tracking for which an SNN calculates optical flow. The entire process, from low-level event-processing to high-level (bit-)sequence-decoding, is schematically depicted in Fig. \ref{fig:architecture}. This figure also introduces specific terms that are used throughout the rest of this paper.

\figArch{b}

The proposed system is a hybrid of a neuromorphic and an algorithmic solution. It follows a major trend in robotics to exploit the rich capabilities of neural networks, which provide sophisticated signal processing and control capabilities \citep{li_distributed_2017}. Simultaneously, to handle the temporal and noisy nature of the real-world signals, neural networks can be extended to handle time delays \citep{jin_neural_2022}, or to include stages with Kalman filtering \citep{yang_highly_2023}, leading to a synergy between neural networks and classic algorithms. Our system follows a similar approach and subsequent paragraphs describe its components.

\subsection{Event-Based Communication}
\label{ssec:protocol}

The emitter is synchronously transmitting, with a blinking pattern, a binary sequence $S$ that consists of a start code $S_c$, a data payload (identification number) $S_p$ and a parity bit $f(S_p)$, where $f$ returns 1 if $S_p$ has an even number of ones, or 0 otherwise. The start code and the parity bit delimit the sequence and confirm its validity, as illustrated in Fig. \ref{fig:protocol}. On the receiver side, the event camera asynchronously generates events upon pixel brightness changes, which can be caused by either a change in the beacon's signal or visual noise in the scene. The current state of the beacon (\textbf{on} or \textbf{off}) cannot be detected by the sensor. Rather, the sensor detects when the beacon transitions between these states. The signal frequency being known, the delay between those transitions gives the number of identical bits emitted. In comparison to a similar architecture with a frame-based camera (200~Hz frame rate) \citep{IV2007}, our setup relies on an event camera and a beacon blinking in kHz frequency, allowing for a short beacon decoding time, better separation from noise and easier tracking since beacon's motions are relatively slower. 

\figSeqSum{b}

As the start code $S_c$ is fixed and the identification number $S_p$ is invariable per beacon, the parity bit $f(S_p)$ remains the same from one sequence to the next. As a result, once the beacon parameters are set, it repeatedly emits the same 11-bit fixed-length frame. The decoding of the transmitted signal exploits these two transmission characteristics. As the cameras do not necessarily pick up the signal exactly from the start code, 11 consecutive bits are stored in memory. If the signal is received correctly, these 11 bits constitute a full sequence. Once this sequence of 11 bits is recovered, it is necessary to search for the subsequence of four bits corresponding to the start code $S_c$ (marked below in bold), which enables to recover a complete sequence through bit rotation:

\begin{itemize}
    \item Reception of 11 successive bits: 0 0 0 0 0 1 \textbf{1 1 1 0} 1
    \item Sequence reconstruction after start code detection: 
\textbf{1 1 1 0} 1 0 0 0 0 0 1
\end{itemize}

\subsection{Object Tracking}

Beacons isolated by the clustering and filtering steps described in Fig. \ref{fig:architecture} are called targets. These are instant detections of the beacons. But these need be tracked in order to extract the blinking code that they produce. The tracked targets are called tracks. They hold a position (estimated or real), the history of state changes (ons and offs) and meta information like a confidence value. Tracks are categorized with types that can change over time. They can be:
    \begin{itemize}
         \item new: the target cannot be associated with any existing track: create a new track
         \item valid: the track's state change history conforms to the communication protocol
         \item invalid: the track's state change history does not conform to the communication protocol (typically noise or continuous signals like solar reflections). Note that a track can change from invalid to valid if it's confidence value rises (detailed later).
    \end{itemize}

\subsubsection{Clustering}
\label{ssec:clustering}
Camera events are being accumulated in a time window and clustered with the Density-Based Spatial Clustering of Applications with Noise \citep{DBSCAN}, chosen to get rid of noisy, isolated events and to retrieve meaningful objects from the visual scene. Such clusters are filtered according to:
\begin{align}
    \frac{N_e}{\pi \times |b-d|^2} > r,\;\; 
    N_e \in \{N_{\text{min}};N_{\text{max}}\}
\end{align}
where $N_e$ is the number of events in the cluster, $|b-d|$ the Euclidean distance between the cluster's barycenter $b$ and $d$ its most distant event, $r$ a shape ratio, and $N_{\text{min}}$ and $N_{\text{max}}$ the minimal/maximal emitter size in pixel. The shape ratio is a hyperparameter. In our setup, it characterizes the roundness of the cluster, since we are looking for round beacons. It can be adapted to other shapes if beacons need be flatter for example. Experimentally, the shape ratio $r$ turned out to play a crucial role in the communication's accuracy: limiting the detection to high ratios (from 0.8 to 0.99) gave the best results. The minimal target size $N_{\text{min}}$ of target must also be carefully set to be able to detect beacons, but small values also imply filtering less noise and having to process more clusters. Depending on the scenario distances, values from 5 to 30 events were chosen.

We reduce the remaining clusters to their barycenter, size and their polarity and call these "targets". The polarity of a target $P$ is given by $P=(\sum_{i}p_i)/ N_e$ where $p_i$=$1$ for a positive polarity and $-1$ for a negative one for each event $i$.

\subsubsection{Event-Based Optical Flow}

Event-based optical flow is calculated by a neural network and processed by the remaining algorithmic beacon tracking pipeline. We introduce it as a given input in the main tracking algorithm presented in the next section.

\figSNN{b}
\label{ssec:of}

Optical flow is computed from the same camera and events that are used for decoding, and delivers a sparse vector field for visible events with velocity and direction.

We implemented an SNN architecture with Spiking Neural Units (SNUs) \citep{wozniak_deep_2020} and extended the model with synaptic delays that we call $\Delta$SNU. Its state equations are:
\begin{align}
s_t &= g(W d_\Delta(x_t) + l(\tau)  s_{t-1} (1-y_{t-1}) ) \\
y_t &= h(s_t - v_{\text{th}}),
\end{align}
where $W$ are the weights, $v_{\text{th}}$ is a threshold, $s_t$ is the state of the neuron and $l(\tau)$ its decay rate, $y_t$ is the output, $g$ is the input activation function, $h$ is the output activation function, and $d$ is the synaptic delay function.
The delay function $d$ is parameterized with a delay matrix $\Delta$ that for each neuron and synapse determines the delay at which spikes from each input $x_t$ will be delivered for the neuronal state calculation.

Optical flow is computed by a CNN with $5\times5$ kernels, illustrated in Fig.\ref{fig:synapticdelays}a.
Each $\Delta$SNU is attuned to the particular direction and speed of movement through its specific synaptic delays, similarly to \citep{orchard_spiking_2013}. When events matching the gradient of synaptic delays are observed, a strong synchronized stimulation of the neuron leads to neuronal firing. This results in sparse detection of optical flow.
The synaptic delay kernels are visualized in Fig. \ref{fig:synapticdelays}b. We use 8 directions and 4 magnitudes, with the maximum delay period corresponding to 10 executions of the tracking algorithm. Weights are set to one and $v_{\text{th}}=5$. The parameters were determined empirically so as to yield the best tracking results. Decreasing the threshold $v_{\text{th}}$ yields faster detection of optical flow, but increases the false positive spikes. Increasing the number of detected directions and magnitudes theoretically provides more accurate estimation of the optical flow. However, in practice it results in false positive activation of neurons detecting similar directions or magnitudes unless $v_{\text{th}}$ is increased at the expense of increased detection latency. 

\subsubsection{Tracking}
\label{ssec:tracking}
Targets are kept in memory for tracking over time and are then called tracks. A Kalman filter is attributed to each track and updated for every processed time window, as depicted in Fig. \ref{fig:tracking}. A Kalman filter is needed to estimate the position of a track from the last measured one and when is not visible, either because of an occlusion, or simply because it transitioned to \textbf{off}. We use the estimated position's optical flow value to draw a search window in which a target is looked for. Similarly to \citep{kalmanfilteropticalflowtracking}, predicted tracks' states are matched to detected targets to minimize the L1-norm between tracks and targets. Unmatched targets are registered as new tracks.

\figTrac{b}

\subsubsection{Identification}
\label{ssec:id}
A matched track's sequence is updated using the target's mean event polarity $P$.
\begin{itemize}
  \item If $P\geq0.5$ then the beacon is assumed to have undergone an \textbf{on} transition. We add $n=(t_c-t_t) / f_{\text{beacon}}$ zeros to the binary sequence where $t_c$ is the current timestamp, $t_t$ is the stored last transition timestamp and $f_{\text{beacon}}$ is the beacon blinking frequency and set $t_t=t_c$.
  \item If $P\leq-0.5$ then the beacon is assumed to have undergone a transition to the \textbf{off} state. Likewise, we add $n$ ones to the binary sequence.
  \item Otherwise, the paired beacon most likely has not undergone a transition but just moved.
\end{itemize}
Similarly to \citep{IV2007}, a confidence value is incremented or decremented to classify tracks as new, valid or invalid, as illustrated in Alg.\ref{alg:confidence}. Indeed, noise can pass clustering filters but will soon be invalidated as its confidence will never rise. To correct for errors (for instance due to occlusions), the confidence increments are larger than decrements. When a track's sequence is long enough to be decoded, it is declared valid if it complies to the protocol and maintains the same payload (if this track was previously correctly recognized). New tracks have an initial confidence value $\leq$ confidence$_{\text{max}}$. These values have been experimentally set to optimize for our protocol and an expected mean occlusion duration. They can be adapted for expected longer \textbf{off} states or longer occlusions. Though, the level of track robustness to occlusion and its "stickyness" have to be balanced. Indeed, higher confidence thresholds lead to a longer detection time and also a longer time to become invalid. A clean up of tracks having been invalid for too long is necessary in all cases to save memory. This is done with a simple threshold (confidence$_{\text{min}}$) or a time out (delay$_{\text{max}}$) mechanism. These hyper-parameters were tunedexperimentally, and we set them to confidence$_{\text{min}}=0$, confidence$_{\text{max}}=20$, and the initial confidence  to  10.

To ensure real-time execution, the tracking occurs at a lower frequency, while the decoding occurs at the emitter frequency. To achieve this, we only accumulate events in the surrounding of existing tracks, and tracks' sequences are updated accordingly.

\figAlgorithm{b}

\subsubsection{Computational Performance}

The hardware event-based camera can detect up to millions of events per second, where many of them may correspond to noise, especially in outdoor and moving camera scenarios. The tracking algorithm, based on a neuronal implementation of optical flow and on a clustering algorithm with a square complexity on the number of events, is computational much more demanding than the decoding algorithm. Therefore, to ensure real-time performance, both loops have been decoupled, so that the tracking is updated at a lower frequency than the decoding, as illustrated in Fig. \ref{fig:computation_reduction}. In this implementation, tracking steps described above occur at 10~Hz, while the decoding happens at up to 5kHz. The relative motion of tracked objects in the visual scene being slow compared to the communication event rate, the tracking update rate is sufficient. To ensure a working communication, the decoding algorithm must be fast and computationally inexpensive to match the emitter frequency.

\figLoops{b}

\section{Results}
\label{sec:res}



\subsection{Static Identification}

Our hardware beacon has four infrared LEDs (850~nm) and an ESP32 micro-controller to set the payload $S_p=42$ and the blinking frequency. A study was conducted to find the optimal wavelenght where the LEDs must be detected as far as possible in an outdoor use case, as described in Fig. \ref{fig:LED}.
\figled{b}
To receive the signal, we used a DVXplorer Mini camera, with a resolution of 640$\times$480 and a 3.6~mm focal length lens. 
In a static indoor setup, the hardware event camera enables us to achieve high data transmission frequencies, plotted in Fig. \ref{fig:hwresultsa}. The metric is the Message Accuracy Rate (MAR): the percentage of correct 11-bit sequences decoded from the beacon's signal during a recording. The MAR stays over 94~\% up to 2.5~kHz, then decreases quickly, due to the temporal resolution of the camera. Using a 16~mm focal length lens we could identify the beacon at a distance of 11.5~m indoors, with 87~\% MAR and a frequency of 1~kHz and obtained 100~\% MAR at 100~Hz at 16~m -- see Fig. \ref{fig:hwresultsb}.

A special note has to be made regarding the range. Results are given here for information purpose. The range cannot really be considered a benchmarking parameter because it depends essentially on the beacon signal power and on the camera lens. To improve the detection and MAR at a longer range, adding LEDs to the beacon, or choosing a zooming lens, are good solutions. So this is basically an implementation choice.

\figResa{b}
\figResb{b}

\subsection{Dynamic Identification}
To evaluate our identification approach in a dynamic setup, where tracking is required, a simulated use case was developed in the Neurorobotics Platform \citep{NRP}. A Hector drone model, with an on-board event camera plugin \citep{event_based_camera_plugin}, flies over a construction site with assets (packages and workers) to be identified and tracked. These are equipped with blinking beacons. The drone follows a predefined trajectory and the scene is captured from a bird's eye view -- see Fig. \ref{fig:simu}. A frame-based camera with the same view is used for visualization. Noise is simulated with beacons of different sizes blinking randomly. For varying drone trajectories, assets were correctly identified at up to 28~m, with drone speeds up to 10~m/s (linear) and 0.5~radian/s (self rotational). Movements were fast, relative to the limited 50~Hz beacon frequency imposed by the simulator. A higher MAR was obtained with a Kalman filter integrating optical flow (\ref{ssec:tracking}) than without it -- see Tab. \ref{tab:comparison}. MAR and Bit Accuracy Rate (BAR) are correlated in simulation because they drop together only upon occlusion. Finally, we conducted hardware experiments where a beacon was moved at 2~m/s reaching a 94~\% BAR at 5m and a 87~\% BAR at 16m. This shows that our system enables accurate identification and data transmission even with moving beacons, which, to our knowledge, is beyond the state-of-the-art.

\figSim{b}

\section{Discussion}
\label{sec:disc}
\tableResults{}

We propose a novel approach for identification that combines the benefits of event-based fast optical communication and signal tracking with spiking optical flow.
The approach was validated in a simulation of drone-based asset monitoring on a construction site. A hardware prototype setup reached state-of-the-art optical communication speed and range. We propose the first -- to the best of our knowledge -- system to identify fast moving, variable beacons with an event camera, thanks to our original tracking approach.
Event-based camera, thanks to their extremely low pixel latency, do outperform OCC based on frame grabbers by orders of magnitude. This enables beacon signal frequencies up to 5 kHz, which in turn, enables for their more robust tracking, since their relative movement is slow between two LED state transitions. Nevertheless, tracking is still necessary for beacons moving fast and that is where this work goes beyond \citep{australianpaperocc}, which assumes null or negligible beacon movement.

Further research includes the port of optical flow computation to neuromorphic hardware and the full port of the system onto a real drone, for real world assessment.
Although the current work is mainly algorithmic with optical flow realized in a spiking neural network, this paper proves that it is very efficient. Now, as this mixes two computing paradigms (algorithmics and spiking neural networks), it will entail having two computing devices on board a real drone.
Another research direction of ours is thus to investigate a full spiking implementation, so as to carry only neuromorphic hardware on board.

\section*{Conflict of Interest Statement}
The authors declare that the research was conducted in the absence of any commercial or financial relationships that could be construed as a potential conflict of interest.

\section*{Funding Statement}
This study was conducted by the participating entities on their own resources. There was no cross-funding or external funding. The commercial entity was not involved in the funding of the other entities' work.

\section*{Author Contributions}

AvA and JL have carried out the research and implementation of the algorithmic and integration work. NEB and SW have investigated and implemented the optical flow neural network, under supervision of AP. AP and AvA have supervised the work for respectively IBM and fortiss and have approved the content of this publication.

\section*{Acknowledgements}
\label{sec:ack}
The research was carried out within the Munich Center for AI (C4AI), a joint entity from fortiss and IBM.
We thank the C4AI project managers for their constructive help.
The research at fortiss was supported by the HBP Neurorobotics Platform funded through the European Union’s Horizon 2020 Framework Program for Research and Innovation under the Specific Grant Agreements No.\,945539 (Human Brain Project SGA3).

\bibliographystyle{Frontiers-Harvard} 
\bibliography{test}

\begin{thebibliography}{21}
\providecommand{\natexlab}[1]{#1}
\expandafter\ifx\csname urlstyle\endcsname\relax
  \providecommand{\doi}[1]{doi:\discretionary{}{}{}#1}\else
  \providecommand{\doi}{doi:\discretionary{}{}{}\begingroup
  \urlstyle{rm}\Url}\fi
\providecommand{\selectlanguage}[1]{\relax}
\providecommand{\bibAnnoteFile}[1]{%
  \IfFileExists{#1}{\begin{quotation}\noindent\textsc{Key:} #1\\
  \textsc{Annotation:}\ \input{#1}\end{quotation}}{}}
\providecommand{\bibAnnote}[2]{%
  \begin{quotation}\noindent\textsc{Key:} #1\\
  \textsc{Annotation:}\ #2\end{quotation}}

\bibitem[{Barranco et~al.(2018)Barranco, Fermuller, and
  Ros}]{tracking_event_based}
Barranco, F., Fermuller, C., and Ros, E. (2018).
\newblock Real-time clustering and multi-target tracking using event-based
  sensors.
\newblock In \emph{2018 IEEE/RSJ International Conference on Intelligent Robots
  and Systems (IROS)}. 5764--5769.
\newblock \doi{10.1109/IROS.2018.8593380}
\bibAnnoteFile{tracking_event_based}

\bibitem[{Cahyadi et~al.(2020)Cahyadi, Chung, Ghassemlooy, and Hassan}]{OCC}
Cahyadi, W.~A., Chung, Y.~H., Ghassemlooy, Z., and Hassan, N.~B. (2020).
\newblock Optical camera communications: Principles, modulations, potential and
  challenges.
\newblock \emph{Electronics} 9.
\newblock \doi{10.3390/electronics9091339}
\bibAnnoteFile{OCC}

\bibitem[{Censi et~al.(2013)Censi, Strubel, Brandli, Delbruck, and
  Scaramuzza}]{ScaramuzzaALMlocalisation}
Censi, A., Strubel, J., Brandli, C., Delbruck, T., and Scaramuzza, D. (2013).
\newblock Low-latency localization by active led markers tracking using a
  dynamic vision sensor.
\newblock In \emph{2013 IEEE/RSJ International Conference on Intelligent Robots
  and Systems}. 891--898.
\newblock \doi{10.1109/IROS.2013.6696456}
\bibAnnoteFile{ScaramuzzaALMlocalisation}

\bibitem[{Chen et~al.(2019)Chen, Zhao, and
  Li}]{kalmanfilteropticalflowtracking}
Chen, Y., Zhao, D., and Li, H. (2019).
\newblock Deep kalman filter with optical flow for multiple object tracking.
\newblock In \emph{2019 IEEE International Conference on Systems, Man and
  Cybernetics (SMC)}. 3036--3041.
\newblock \doi{10.1109/SMC.2019.8914078}
\bibAnnoteFile{kalmanfilteropticalflowtracking}

\bibitem[{Du et~al.(2013)Du, Ibrahim, Shehata, and Badawy}]{licenseplate}
Du, S., Ibrahim, M., Shehata, M., and Badawy, W. (2013).
\newblock Automatic license plate recognition (alpr): A state-of-the-art
  review.
\newblock \emph{IEEE Transactions on Circuits and Systems for Video Technology}
  23, 311--325.
\newblock \doi{10.1109/TCSVT.2012.2203741}
\bibAnnoteFile{licenseplate}

\bibitem[{Falotico et~al.(2017)Falotico, Vannucci, Ambrosano, Albanese,
  Ulbrich, Vasquez~Tieck et~al.}]{NRP}
Falotico, E., Vannucci, L., Ambrosano, A., Albanese, U., Ulbrich, S.,
  Vasquez~Tieck, J.~C., et~al. (2017).
\newblock Connecting artificial brains to robots in a comprehensive simulation
  framework: The neurorobotics platform.
\newblock \emph{Frontiers in Neurorobotics} 11.
\newblock \doi{10.3389/fnbot.2017.00002}
\bibAnnoteFile{NRP}

\bibitem[{Gehrig et~al.(2021)Gehrig, Millhäusler, Gehrig, and
  Scaramuzza}]{gehrig_e-raft_2021}
Gehrig, M., Millhäusler, M., Gehrig, D., and Scaramuzza, D. (2021).
\newblock E-{RAFT}: {Dense} {Optical} {Flow} from {Event} {Cameras}.
\newblock \emph{arXiv:2108.10552 [cs]} ArXiv: 2108.10552
\bibAnnoteFile{gehrig_e-raft_2021}

\bibitem[{ITU(2006)}]{UWB}
ITU (2006).
\newblock Characteristics of ultra-wideband technology
\bibAnnoteFile{UWB}

\bibitem[{Jia et~al.(2012)Jia, Feng, Fan, and Lei}]{rfid}
Jia, X., Feng, Q., Fan, T., and Lei, Q. (2012).
\newblock Rfid technology and its applications in internet of things (iot).
\newblock In \emph{2012 2nd International Conference on Consumer Electronics,
  Communications and Networks (CECNet)}. 1282--1285.
\newblock \doi{10.1109/CECNet.2012.6201508}
\bibAnnoteFile{rfid}

\bibitem[{Jin et~al.(2022)Jin, Zheng, and Luo}]{jin_neural_2022}
Jin, L., Zheng, X., and Luo, X. (2022).
\newblock Neural dynamics for distributed collaborative control of manipulators
  with time delays.
\newblock \emph{IEEE/CAA Journal of Automatica Sinica} 9, 854--863.
\newblock \doi{10.1109/JAS.2022.105446}
\bibAnnoteFile{jin_neural_2022}

\bibitem[{Kaiser et~al.(2016)Kaiser, Tieck, Hubschneider, Wolf, Weber, Hoff
  et~al.}]{event_based_camera_plugin}
Kaiser, J., Tieck, J. C.~V., Hubschneider, C., Wolf, P., Weber, M., Hoff, M.,
  et~al. (2016).
\newblock Towards a framework for end-to-end control of a simulated vehicle
  with spiking neural networks.
\newblock In \emph{2016 IEEE International Conference on Simulation, Modeling,
  and Programming for Autonomous Robots (SIMPAR)}. 127--134.
\newblock \doi{10.1109/SIMPAR.2016.7862386}
\bibAnnoteFile{event_based_camera_plugin}

\bibitem[{Li et~al.(2017)Li, He, Li, and Rafique}]{li_distributed_2017}
Li, S., He, J., Li, Y., and Rafique, M.~U. (2017).
\newblock Neural dynamics for distributed collaborative control of manipulators
  with time delays.
\newblock \emph{IEEE Transactions on Neural Networks and Learning Systems} 28,
  415--426.
\newblock \doi{10.1109/TNNLS.2016.2516565}
\bibAnnoteFile{li_distributed_2017}

\bibitem[{Ojeda et~al.(2020)Ojeda, Bisulco, Kepple, Isler, and
  Lee}]{pedestriandetection_eventbased}
Ojeda, F.~C., Bisulco, A., Kepple, D., Isler, V., and Lee, D.~D. (2020).
\newblock On-device event filtering with binary neural networks for pedestrian
  detection using neuromorphic vision sensors.
\newblock In \emph{2020 IEEE International Conference on Image Processing
  (ICIP)}. 3084--3088.
\newblock \doi{10.1109/ICIP40778.2020.9191148}
\bibAnnoteFile{pedestriandetection_eventbased}

\bibitem[{Orchard et~al.(2013)Orchard, Benosman, Etienne-Cummings, and
  Thakor}]{orchard_spiking_2013}
Orchard, G., Benosman, R., Etienne-Cummings, R., and Thakor, N.~V. (2013).
\newblock A spiking neural network architecture for visual motion estimation.
\newblock In \emph{2013 {IEEE} {Biomedical} {Circuits} and {Systems}
  {Conference} ({BioCAS})} (Rotterdam, Netherlands: IEEE), 298--301.
\newblock \doi{10.1109/BioCAS.2013.6679698}
\bibAnnoteFile{orchard_spiking_2013}

\bibitem[{Perez-Ramirez et~al.(2019)Perez-Ramirez, Roberts, Navik,
  Muralidharan, and Moustafa}]{binningproblemeventbased}
Perez-Ramirez, J., Roberts, R.~D., Navik, A.~P., Muralidharan, N., and
  Moustafa, H. (2019).
\newblock Optical wireless camera communications using neuromorphic vision
  sensors.
\newblock In \emph{2019 IEEE International Conference on Communications
  Workshops (ICC Workshops)}. 1--6.
\newblock \doi{10.1109/ICCW.2019.8756795}
\bibAnnoteFile{binningproblemeventbased}

\bibitem[{Schnider et~al.(2023)Schnider, Woźniak, Gehrig, Lecomte, von Arnim,
  Benini et~al.}]{schnider_2023}
Schnider, Y., Woźniak, S., Gehrig, M., Lecomte, J., von Arnim, A., Benini, L.,
  et~al. (2023).
\newblock Neuromorphic optical flow and real-time implementation with event
  cameras.
\newblock \emph{IEEE Conference on Computer Vision and Pattern Recognition
  Workshops (CVPRW), Vancouver, 2023. c IEEE}
\bibAnnoteFile{schnider_2023}

\bibitem[{Teed and Deng(2020)}]{teed_raft_2020}
Teed, Z. and Deng, J. (2020).
\newblock {RAFT}: {Recurrent} {All}-{Pairs} {Field} {Transforms} for {Optical}
  {Flow} ArXiv:2003.12039 [cs]
\bibAnnoteFile{teed_raft_2020}

\bibitem[{vonArnim et~al.(2007)vonArnim, Perrollaz, Bertrand, and
  Ehrlich}]{IV2007}
vonArnim, A., Perrollaz, M., Bertrand, A., and Ehrlich, J. (2007).
\newblock Vehicle identification using near infrared vision and applications to
  cooperative perception.
\newblock In \emph{2007 IEEE Intelligent Vehicles Symposium}. 290--295.
\newblock \doi{10.1109/IVS.2007.4290129}
\bibAnnoteFile{IV2007}

\bibitem[{Wang et~al.(2022)Wang, Ng, Henderson, and
  Mahony}]{australianpaperocc}
Wang, Z., Ng, Y., Henderson, J., and Mahony, R. (2022).
\newblock Smart visual beacons with asynchronous optical communications using
  event cameras.
\newblock In \emph{2022 IEEE/RSJ International Conference on Intelligent Robots
  and Systems (IROS)}. 3793--3799.
\newblock \doi{10.1109/IROS47612.2022.9982016}
\bibAnnoteFile{australianpaperocc}

\bibitem[{Woźniak et~al.(2020)Woźniak, Pantazi, Bohnstingl, and
  Eleftheriou}]{wozniak_deep_2020}
Woźniak, S., Pantazi, A., Bohnstingl, T., and Eleftheriou, E. (2020).
\newblock Deep learning incorporating biologically inspired neural dynamics and
  in-memory computing.
\newblock \emph{Nature Machine Intelligence} 2, 325--336.
\newblock \doi{10.1038/s42256-020-0187-0}
\bibAnnoteFile{wozniak_deep_2020}

\bibitem[{Yang et~al.(2023)Yang, Li, Li, and Luo}]{yang_highly_2023}
Yang, W., Li, S., Li, Z., and Luo, X. (2023).
\newblock Highly accurate manipulator calibration via extended {Kalman}
  filter-incorporated residual neural network.
\newblock \emph{IEEE Transactions on Industrial Informatics} 19, 10831--10841.
\newblock \doi{10.1109/TII.2023.3241614}
\bibAnnoteFile{yang_highly_2023}

\end{thebibliography}


\end{document}